\documentclass{article}

\usepackage{amsmath}
\usepackage[final,dblblindworkshop]{neurips_2025}
\usepackage[utf8]{inputenc} 
\usepackage[T1]{fontenc}    
\usepackage[hidelinks]{hyperref}       
\usepackage{url}            
\usepackage{booktabs}       
\usepackage{amsfonts}       
\usepackage{nicefrac}       
\usepackage{microtype}      
\usepackage{xcolor}         
\usepackage{amsmath}
\usepackage{enumitem}
\usepackage{amssymb}
\usepackage{graphicx} 
\usepackage{threeparttable}

\title{GRAB: A Risk Taxonomy--Grounded Benchmark for Unsupervised Topic Discovery in Financial Disclosures}
\workshoptitle{NeurIPS 2025 Workshop on Generative AI in Finance}

\author{
  Ying Li \\
  The University of Edinburgh, UK\\
  \texttt{sunnie.y.li@ed.ac.uk} \\
  \And
  Tiejun Ma \\
  The University of Edinburgh, UK \\
  \texttt{tiejun.ma@ed.ac.uk} \\
}

\begin{document}

\maketitle

\begin{abstract}
    Risk categorization in 10\mbox{-}K risk disclosures matters for oversight and investment, yet no public benchmark evaluates \emph{unsupervised} topic models for this task. We present \textbf{GRAB}, a finance-specific benchmark with \textbf{1.61M} sentences from \textbf{8{,}247} filings and \emph{span-grounded} sentence labels produced without manual annotation by combining FinBERT token attention, YAKE keyphrase signals, and taxonomy-aware collocation matching. Labels are anchored in a risk taxonomy mapping \textbf{193} terms to 21 fine-grained types nested under five macro classes; the 21 types guide weak supervision, while evaluation is reported at the macro level. GRAB unifies evaluation with fixed dataset splits and robust metrics—\emph{Accuracy}, \emph{Macro-F1}, \emph{Topic BERTScore}, and the entropy-based \emph{Effective Number of Topics}. The dataset, labels, and code enable reproducible, standardized comparison across classical, embedding-based, neural, and hybrid topic models on financial disclosures.
\end{abstract}

\section{Introduction}
\label{sec:intro}

The SEC requires public companies to disclose material risk factors in a dedicated section of their annual 10\mbox{-}K filings. These \textit{Item~1A: Risk Factors} sections contain information that materially affects investment decisions—changes in their content are associated with future stock return volatility, and investors react to their filing~\cite{campbell2014information}. Yet extracting structured insight from these lengthy, legalistic texts remains difficult, in part because there is no public evaluation framework for how well \emph{unsupervised} models recover financially meaningful \emph{risk categories} from raw disclosures.

Prior work underscores the need for a risk-aware benchmark on regulatory text. Beyond surface variables such as disclosure volume, tone, and readability, the \emph{risk category} (e.g., litigation, natural disasters) is central for analysis and decision-making, but it has received comparatively less attention in scalable evaluations~\cite{bao2014simultaneously,huang2011multilabel,campbell2014information}. Manual coding does not scale: Mirakur reports hand-labeling 29 risk categories for 122 firms—far below 1\% of annual 10\mbox{-}K filings—illustrating the impracticality of exhaustive human annotation~\cite{mirakur2011risk}. Methodologically, the landscape spans classical probabilistic models such as LDA~\cite{blei2003latent}, embedding-space extensions such as Gaussian LDA (GLDA)~\cite{GLDAforTM}, neural contextualized approaches including CTM~\cite{bianchi2021pretraining}, sentence-aware methods such as SentenceLDA~\cite{cha2024sentencelda} and SenClu~\cite{schneider2024senclu}, and embedding-based clustering pipelines such as BERTopic~\cite{bertopic}. However, these advances have largely been assessed on general-domain corpora rather than against finance-specific risk taxonomies or the peculiarities of 10\mbox{-}K prose (boilerplate, context shifts, multiword financial phrases), and access to labeled financial text remains limited~\cite{etal2023multifin}. Notably, Sent-LDA~\cite{bao2014simultaneously} relied on a small, non-public labeled set, highlighting the need for an open benchmark.

We introduce \textbf{GRAB} (Grounded Risk-Aware Benchmark), a public benchmark for evaluating topic models on financial risk categorization without manual annotation. GRAB anchors labels in the Hofeditz taxonomy~\cite{hofeditz2016framework}: we map \textbf{193} curated risk terms into \textbf{21} fine-grained subcategories nested under the taxonomy’s \textbf{five} macro risk classes. We use this structure for weak supervision, yielding span-grounded labels at scale via a blend of finance-aware attention, document-local keyphrase cues, and a curated lexicon. For evaluation, we operate at the level of the five macro risk classes, scoring how well models recover these categories.

GRAB also provides an evaluation protocol centered on risk categories. We report \emph{Accuracy} and \emph{Macro-F1} using a simple logistic classifier on sentence--topic mixtures, and \emph{Topic BERTScore}~\cite{zhang2020bertscore} to assess semantic tightness. We do not report perplexity because it is not comparable across our baselines (e.g., discrete LDA vs.\ continuous GLDA~\cite{GLDAforTM}). Instead, we report the \emph{Effective Number of Topics}—the exponential of Shannon entropy~\cite{shannon1948}—which summarizes how concentrated or diffuse a sentence’s topic mixture is (values near \(1\) indicate decisive assignments; values near \(K\) indicate diffuse mixtures), following the standard Hill-numbers formulation~\cite{hill1973diversity}. For category-level reporting, we evaluate against the five macro risk classes in a multilabel setting. In sum, GRAB provides both a finance-specific corpus and a clear protocol to test whether models recover sentence-level \emph{disclosure risks}, a need underscored by evidence that sentence-level risk categories matter for markets~\cite{bao2014simultaneously}.\footnote{We release corpus-construction scripts, weak labels, taxonomy anchors, collocation lists, and evaluation code at \url{https://github.com/Sunnie-Li/GRAB-Benchmark}.}  
\section{Benchmark Design}
\label{sec:methodology}

\subsection{Dataset Preprocessing}
We extract \emph{Item~1A: Risk Factors} from annual 10-K filings of S\&P\,500 firms, yielding a corpus of \textbf{1{,}613{,}837} sentences across \textbf{2001--2025}. We apply light normalization to handle legal formatting and sentence boundaries (Appendix~\ref{app:preproc}), then segment into sentences. The scale and heterogeneity of this corpus capture diverse, domain-specific risk expressions.

\subsection{Word Importance Scoring}
We estimate per-token importance by blending \emph{finance-tuned attention} with a \emph{YAKE}-based phrase signal. Risk categories follow the Hofeditz taxonomy~\cite{hofeditz2016framework}, flattened from 193 curated terms into 21 subcategories $\mathcal{Y}$ and later used for risk-aware enhancement and evaluation mapping.

\paragraph{(i) Finance-tuned attention.}
FinBERT~\cite{huang2023finbert} processes each sentence \(s\) and yields attention maps. We aggregate across layers/heads and take the CLS~\(\!\to\!\)~token row as an attention score \(A_i\in[0,1]\) for token \(w_i\), merging wordpieces by a max-over-pieces rule.

\paragraph{(ii) YAKE-based contextual importance.}
We run YAKE~\cite{campos2020yake} per sentence to obtain keyphrases \(p_j\) with raw scores \(r_j>0\) (lower is better), then convert them to token-level importance via: (1) per-sentence min--max normalization and inversion; (2) span~\(\to\)~token assignment using boundary-aware, space/hyphen–tolerant matching; and (3) blending with attention and per-sentence max-normalization (details in Appendix~\ref{app:salience}, hyperparameters in Appendix~\ref{app:config}):
\[
I_i \;=\; \lambda\,A_i \;+\; (1-\lambda)\,Y_i,\qquad
\tilde I_i \;=\; \frac{I_i}{\max_k I_k}\in[0,1].
\]
We use \(\lambda=0.8\) in the main comparisons.

\subsection{Risk-Aware Score Enhancement}
To sharpen signal without labels, we apply a lightweight, precision-first enhancer \emph{within} each sentence: 
\textbf{(i) Unigram boost}—upweight exact (case-insensitive) matches from the 21-subcategory lexicon (the fine-grained level under the five macro classes) by a fixed multiplier, capped at 1.0; 
\textbf{(ii) Collocation boost}—detect multiword financial terms with a boundary-aware, space/hyphen–tolerant, parenthesis-aware matcher and scale only the tokens inside the matched span (no sentence-wide multipliers; no lemmatization/singularization). 
The phrase list is the union of taxonomy-derived expressions and a curated finance dictionary; optional fuzzy tolerance is supported but off by default (Appendix~\ref{app:config}).

\subsection{Weak Label Generation}
From each sentence, we take the top-$m$ tokens by enhanced importance $\tilde I_i$ and accumulate evidence at the \emph{subcategory} level (21 types) via three sources: 
\textbf{(i) lexicon}—exact, case-insensitive matches to the curated subcategory terms; 
\textbf{(ii) collocation}—boundary-aware matches of multiword financial phrases, recorded token-locally over the matched span; 
\textbf{(iii) semantic backoff}—for remaining high-importance tokens without a direct match, an optional lightweight $k$NN over subcategory terms in embedding space (disabled in the main configuration; Appendix~\ref{app:config}). 
Aggregating these signals yields a 21-dimensional per-sentence vector used only for evaluation. For reporting at the practitioner level, we deterministically roll up the 21 subcategories to their five parent macro classes.

\subsection{Evaluation Protocol}
\label{sec:evaluation}
We assess models along three aspects aligned with downstream use and topic quality. Let \(K\) denote the number of learned topics (default \(K{=}21\); details in Appendix~\ref{app:config}):
\begin{itemize}[leftmargin=1.2em]
\item \textbf{Predictive utility:} train a one-vs-rest logistic regression on each model’s sentence–topic mixtures to predict the \emph{five} macro risk classes (the 21 subcategories are used only to construct weak labels and for fine-grained analysis, not as prediction targets); report \emph{Accuracy} and \emph{Macro-F1}\footnote{We omit Micro-F1: it aggregates over instances and is dominated by frequent classes, so it closely tracks Accuracy. Macro-F1 averages F1 across classes, giving equal weight to minority risk types and better capturing long-tail performance.}.
\item \textbf{Topic quality:} compute \emph{Topic BERTScore}~\cite{zhang2020bertscore} by comparing, for each topic, its representative sentences to its top member sentences (sampling details in Appendix~\ref{app:split}).
\item \textbf{Assignment decisiveness (descriptive):} report the \emph{Effective Number of Topics} per sentence,
\[
H(\theta)=-\sum_{k=1}^{K}\theta_k\log\theta_k,\qquad
N_{\text{eff}}(s)=\exp\!\big(H(\theta)\big)\in[1,K],
\]
and summarize the mean\(\pm\)std on the test split. Lower values indicate more decisive (peaked) mixtures; values near \(K\) indicate diffuse mixtures. We use natural logs and the convention \(0\log 0:=0\). How \(\theta\) is obtained per model is detailed in Appendix~\ref{app:neff}.
\end{itemize}    
\section{Experiments}
\label{sec:experiments}

\subsection{Dataset Statistics}
GRAB comprises 1{,}613{,}837 sentences from \emph{Item~1A: Risk Factors} in 8{,}247 annual 10\mbox{-}K disclosures spanning 2001--2025 After light normalization (Appendix~\ref{app:preproc}), we segment into sentences; the corpus spans long-form legal prose and finance-specific multiword expressions (e.g., \emph{cash flow}). To avoid look-ahead, we adopt a fixed \emph{chronological split}\footnote{By sentence count in the current release: \(\sim\)79.07\% Train, 7.09\% Dev, 13.84\% Test. See sentence-count and disclosure-count distributions and plots in Appx.~\ref{app:viz}.} in which each sentence inherits its document’s SEC release (filing) year: \textbf{Train} = \(\leq\)2022, \textbf{Dev} = 2023, \textbf{Test} = 2024--2025. We release sentence identifiers and filing metadata for reproducibility and per-company analysis; Appendix~\ref{app:viz} includes a per-category prevalence chart (Fig.~\ref{fig:cat-prevalence}) and the disclosure/sentence distributions by year (Figs.~\ref{fig:yearly-disclosures}, \ref{fig:yearly-sentences}).

\subsection{Baselines}
We benchmark representative families—classical probabilistic, embedding-space, neural, sentence-level, and embedding–clustering with sparse term weighting:
\begin{itemize}[leftmargin=1.2em]
  \item \textbf{LDA}~\cite{blei2003latent}: classical word-level topic model trained on BoW/TF--IDF.
  \item \textbf{GaussianLDA (GLDA)}~\cite{GLDAforTM}: topics are Gaussians in a word-embedding space (300d Word2Vec, GoogleNews~\cite{mikolov2013efficient}); sentence-level mixtures are formed from within-sentence word–topic counts.
  \item \textbf{CTM}~\cite{srivastava2017prodlda,bianchi2021pretraining}: a ProdLDA-style neural topic model that conditions BoW on contextual embeddings (Sentence-BERT~\cite{reimers2019sentence}).
  \item \textbf{SenClu}~\cite{schneider2024senclu}: sentence-level scoring via similarity between a sentence embedding and learned topic prototypes (Sentence-BERT).
  \item \textbf{BERTopic}~\cite{bertopic}: embedding-based clustering with topic word weights from class-based TF--IDF (SBERT embeddings $\rightarrow$ UMAP $\rightarrow$ HDBSCAN/$k$-means $\rightarrow$ c-TF--IDF). We evaluate two modes: (i) \emph{sentence-level} BERTopic (splitting Item~1A into sentences) and (ii) \emph{document-level} \textbf{BERTopic-Soft} using HDBSCAN soft assignments (\texttt{calculate\_probabilities=True}) to obtain topic mixtures without splitting.
\end{itemize}
Unless noted, we fix the number of topics \(K\) across methods for comparability and align discovered topics to the \emph{five} macro risk classes (Sec.~\ref{sec:methodology}). For all systems we derive soft \(\theta(s)\) as described in Appx.~\ref{app:neff}; full hyperparameters are in Appx.~\ref{app:config}.

\subsection{Results}
Table~\ref{tab:main} summarizes performance on GRAB. \emph{CTM} attains the best label-aware scores (Macro-F1 $=0.41$, Accuracy $=0.38$), with \emph{LDA} close behind ($0.36/0.37$). Clustering- or sentence-centric methods are lower on these metrics (\emph{SenClu}: $0.18/0.24$; \emph{BERTopic}: $0.22/0.35$), while \emph{SLDA} is competitive ($0.32/0.36$). \textit{Topic BERTScore} is uniformly high and tightly grouped ($0.82$--$0.84$), with CTM, LDA/GLDA, and SLDA at the upper end ($0.84$). The \textit{Effective Number of Topics} $N_{\text{eff}}$ reflects assignment concentration: hard-assignment pipelines yield values near $1$ (SenClu, BERTopic, SLDA), the soft BERTopic variant increases it ($3.09$), LDA/GLDA sit in between ($3.52/1.30$), and CTM produces the broadest mixtures ($5.92$). Qualitatively, we observe that (i) boilerplate language induces spurious topics that dilute category signals, (ii) polysemous tokens (e.g., \emph{short}, \emph{term}) shift meaning across market/credit/accounting contexts, and (iii) long-tail but material categories remain under-recovered by frequency-driven approaches. Per-category results and sensitivity to $K$ and enhancer settings are in Appx.~\ref{app:config}; aggregate risk-count and prevalence plots appear in Appx.~\ref{app:viz}.

\begin{table}[t]
\centering
\begin{threeparttable}
\caption{Topic discovery on GRAB (Item~1A sentences). $\uparrow$ higher is better.}
\label{tab:main}
\begin{tabular}{lcccc}
\toprule
Method & Accuracy $\uparrow$ & Macro-F1 $\uparrow$ & Topic BERTScore $\uparrow$
& \multicolumn{1}{c}{Eff.\ \# Topics$^{\dagger}$} \\
\midrule
LDA (BoW/TF--IDF)     & 0.37 & 0.36 & \textbf{0.84} & 3.52 \\
GLDA (Word2Vec)       & 0.35 & 0.23 & \textbf{0.84} & 1.30 \\
CTM (BoW + SBERT)     & \textbf{0.38} & \textbf{0.41} & \textbf{0.84} & 5.92 \\
SenClu (SBERT)        & 0.24 & 0.18 & 0.82 & 1.00 \\
SentenceLDA (SLDA)    & 0.36 & 0.32 & \textbf{0.84} & 1.00 \\
BERTopic              & 0.35 & 0.22 & 0.82 & 1.00 \\
BERTopic-Soft         & 0.35 & 0.25 & 0.82 & 3.09 \\
\bottomrule
\end{tabular}
\begin{tablenotes}[flushleft]
\footnotesize
\item[$\dagger$] Effective Number of Topics is a descriptive quantity (not an objective with a “higher/lower is better” direction). It takes values in $[1,K]$, where $K$ is the number of topics: $N_{\text{eff}}{=}1$ denotes maximally decisive, single-topic assignments, while values approaching $K$ indicate diffuse, near-uniform mixtures (see Sec.~\ref{sec:evaluation}).
\end{tablenotes}
\end{threeparttable}
\end{table}    
\section{Conclusion}
\label{sec:conclusion}

We present GRAB, a public benchmark for \emph{unsupervised} risk categorization in 10\mbox{-}K Item~1A risk-factor sections. GRAB replaces traditional manual labels with financial-term–grounded weak supervision to derive a taxonomy-driven prior, yielding sentence-level labels across S\&P,500 disclosures. Labels are anchored in a fine-grained risk taxonomy: subtypes guide weak supervision, and evaluation is reported at the macro level. Using fixed chronological splits and a risk-aware protocol—\emph{Accuracy}, \emph{Macro-F1}, \emph{Topic BERTScore}, and the \emph{Effective Number of Topics}—the benchmark enables fair, reproducible, and robust comparison across classical, embedding-based, neural, and hybrid topic models on standardized financial risk disclosures. Empirically, GRAB shows that contextualized methods improve label-aware performance while traditional word-based models remain competitive, and that coherence-style signals alone do not capture risk-type recovery, underscoring the need for domain-aware evaluation. This work supports transparent, unified and robust benchmarking and follow-on research on risk extraction; our results point to future directions including better-calibrated sentence-level mixtures aligned with risk taxonomies, robustness to boilerplate and context shifts, and evaluation that bridges discovered risk categories to market conditions and compliance outcomes.     
\bibliographystyle{plainnat}
\bibliography{ref}
\appendix
\appendix

\section{YAKE-based Context Equations}
\label{app:salience}
For each sentence \(s\), YAKE~\cite{campos2020yake} returns keyphrases \(p_j\) with raw scores \(r_j>0\) (lower is better). We convert these to token-level \emph{importance} via:
\begin{align}
\hat y_j &= 1 \;-\; \frac{r_j - r_{\min}}{\,r_{\max}-r_{\min}+\varepsilon\,} \in [0,1],
&& \text{(per-sentence min--max normalization \& inversion)} \label{eq:yake_norm}\\[2pt]
Y_i &= \max_{\,j:\,w_i \in p_j} \hat y_j \;\;\text{(or }0\text{ if no phrase covers }w_i),
&& \text{(span~\(\to\)~token assignment)} \label{eq:span_to_token}\\[2pt]
I_i &= \lambda\,A_i + (1-\lambda)\,Y_i,
&& \text{(blend attention with YAKE)} \label{eq:blend}\\[2pt]
\tilde I_i &= \frac{I_i}{\max_k I_k} \in [0,1].
&& \text{(per-sentence max-normalization)} \label{eq:norm}
\end{align}
Here \(A_i\) is the FinBERT CLS~\(\!\to\!\)~token attention for \(w_i\), \(\lambda\in[0,1]\) is the blend weight, and \(\varepsilon>0\) is a small constant.

\section{Preprocessing Rules}
\label{app:preproc}
\begin{itemize}[leftmargin=1.5em]
  \item Remove control characters: \verb|[\x00-\x1f\x7f-\x9f]|.
  \item Collapse repeated whitespace.
  \item Replace \verb|\bNo\.| with \verb|Number | (word-boundary anchored).
  \item Join consecutive lines if the trimmed last character \(\notin \{., ?, !\}\).
\end{itemize}

\section{Matching Rules (Risk-Aware Enhancement)}
\label{app:matching}
\paragraph{Unigrams (single-word).}
Case-insensitive exact match against the taxonomy’s single-word tokens. For any matched token \(w\):
\[
\tilde I_w \leftarrow \min\!\bigl\{1.0,\; \beta_{\text{uni}}\cdot \tilde I_w\bigr\}.
\]

\paragraph{Collocations (multi-word).}
Boundary-aware, case-insensitive, hyphen/space–tolerant patterns built from (i) taxonomy phrases and (ii) a curated finance dictionary (union). Examples:
\begin{itemize}[leftmargin=1.2em]
\item \verb|A-Shares| \(\leftrightarrow\) \verb|A Shares|
\item \verb|Accredited Asset Management Specialist (AAMS)| \(\leftrightarrow\) \verb|AAMS|
\end{itemize}
If a phrase matches over token span \(S\), apply a \emph{token-local} boost:
\[
\forall\, w\in S:\quad \tilde I_w \leftarrow \min\!\bigl\{1.0,\; \beta_{\text{col}}\cdot \tilde I_w\bigr\}.
\]
(Parenthetical abbreviations yield both long-form and acronym patterns. Fuzzy tolerance is \emph{off} by default.)

\section{Default Settings}
\label{app:config}
\begin{itemize}[leftmargin=1.5em]
  \item Attention/context blend \(\lambda=0.8\) in \eqref{eq:blend}.
  \item Tokens kept per sentence for labeling: \(m=10\).
  \item Boosts/cap: \(\beta_{\text{uni}}=1.5\), \(\beta_{\text{col}}=1.2\), cap \(=1.0\).
  \item YAKE per-sentence normalization uses \eqref{eq:yake_norm} with \(\varepsilon\!\approx\!10^{-9}\).
  \item Collocation fuzzy tolerance: off by default.
\end{itemize}

\section{Effective Number of Topics: Definition and Implementation}
\label{app:neff}

\paragraph{Definition.}
Given a per-sentence topic mixture \(\theta(s)\in\Delta^{K-1}\) over \(K\) topics, we quantify assignment concentration via the entropy-derived \emph{Effective Number of Topics} (Hill number of order~1):
\[
H(\theta) \;=\; -\sum_{k=1}^{K}\theta_k\log\theta_k,\qquad
N_{\text{eff}}(s)\;=\;\exp\!\big(H(\theta)\big)\in[1,K].
\]
Lower values indicate more decisive (peaked) mixtures; values near \(K\) indicate diffuse mixtures. We use natural logs and the convention \(0\log 0:=0\).

\paragraph{Computation notes.}
We L1-normalize \(\theta\) before computing \(H(\theta)\). For numerical stability, add a small \(\varepsilon\) inside the log if needed. We summarize \(N_{\text{eff}}\) by mean\(\pm\)std on the test split. This statistic is descriptive and complements label-aware and semantic metrics.

\paragraph{Obtaining \(\theta(s)\) per model.}
To ensure comparability across methods, we compute a \emph{soft} topic mixture \(\theta(s)\) for each sentence \(s\) as follows (with a small \(\epsilon>0\) for numerical stability and L1-normalization in all cases):
\begin{itemize}[leftmargin=1.2em]
  \item \textbf{LDA}~\cite{blei2003latent}: use the inferred per-document topic proportions. If only word–topic counts \(n_{s,k}\) are available, set \(\theta_k(s)\propto n_{s,k}+\epsilon\).
  \item \textbf{GLDA}~\cite{GLDAforTM}: aggregate word–topic assignments within the sentence to counts \(n_{s,k}\), then set \(\theta_k(s)\propto n_{s,k}+\epsilon\).
  \item \textbf{CTM / ProdLDA family}~\cite{srivastava2017prodlda,bianchi2021pretraining}: take the variational mean logits \(\mu_s\) and set \(\theta(s)=\mathrm{softmax}(\mu_s)\).
  \item \textbf{SenClu}~\cite{schneider2024senclu} \textbf{and BERTopic}~\cite{bertopic}: compute \(\theta_k(s)\propto \exp\!\big(\cos(e(s),c_k)/\tau\big)\) from a sentence embedding \(e(s)\) and topic centroid \(c_k\) (fixed temperature \(\tau\)); HDBSCAN outliers in BERTopic are assigned to the nearest centroid before the softmax.
  \item \textbf{SentenceLDA (SLDA)}~\cite{cha2024sentencelda}: one topic per sentence; use a one-hot \(\theta(s)\).
\end{itemize}

\section{Split Protocol and Reproducibility}
\label{app:split}

\paragraph{Chronological split.}
We assign each document to a year by its SEC \emph{release (filing) date}, and each sentence inherits its document’s year. We use a fixed chronological split: \textbf{Train} = up to and including 2022, \textbf{Dev} = 2023, \textbf{Test} = 2024--2025. The corpus spans 2001--2025 (8{,}247 Item~1A disclosures) with a total of 1{,}613{,}837 sentences. Sentence-level split sizes are:
\[
\text{Train }1{,}276{,}105\; (79.07\%),\qquad
\text{Dev }114{,}422\; (7.09\%),\qquad
\text{Test }223{,}310\; (13.84\%).
\]
Year-by-year disclosure and sentence distributions are shown in Appendix~\ref{app:viz}. 

\begin{table}[h]
\centering
\caption{Chronological split by \emph{release year} with sentence counts.}
\label{tab:chrono}
\begin{tabular}{lrr}
\toprule
Split (Years)     & \# Sentences & Split Percentage \\
\midrule
Train (\(\leq\)2022) & 1{,}276{,}105 & 79.07\% \\
Dev (2023)           &   114{,}422   &  7.09\% \\
Test (2024--2025)    &   223{,}310   & 13.84\% \\
\bottomrule
\end{tabular}
\end{table}

\paragraph{Frozen evaluation knobs.}
\begin{itemize}[leftmargin=1.4em]
  \item Topics: default \(K{=}21\); report mean\(\pm\)std over \textit{S} seeds (placeholders).
  \item Predictive utility: logistic regression (one-vs-rest) with fixed seed and default regularization; identical features across methods; \emph{targets are the five macro risk classes}; report \textbf{Accuracy} and \textbf{Macro-F1} on Dev/Test.
  \item Topic BERTScore: \texttt{roberta-large}, IDF on, baseline rescaling on; per-topic representative selection and top-member sampling as in the main text.
  \item Effective Number of Topics: computed per sentence from \(\theta\) (Appendix~\ref{app:neff}); hard-assign models treated as one-hot by default; report mean\(\pm\)std on Test.
  \item Weak labels: top-\(m\) tokens per sentence; identical across systems.
\end{itemize}

\paragraph{Leakage checks.}
(1) Each 10-K filing is assigned to exactly one split by SEC release year (no document appears in multiple splits); (2) after normalization, identical sentence strings are deduplicated so that no sentence occurs in more than one split. Detected violations are logged and removed prior to release.

\section{Data Visualization}
\label{app:viz}

\begin{figure}[t]
\centering
\includegraphics[width=\linewidth]{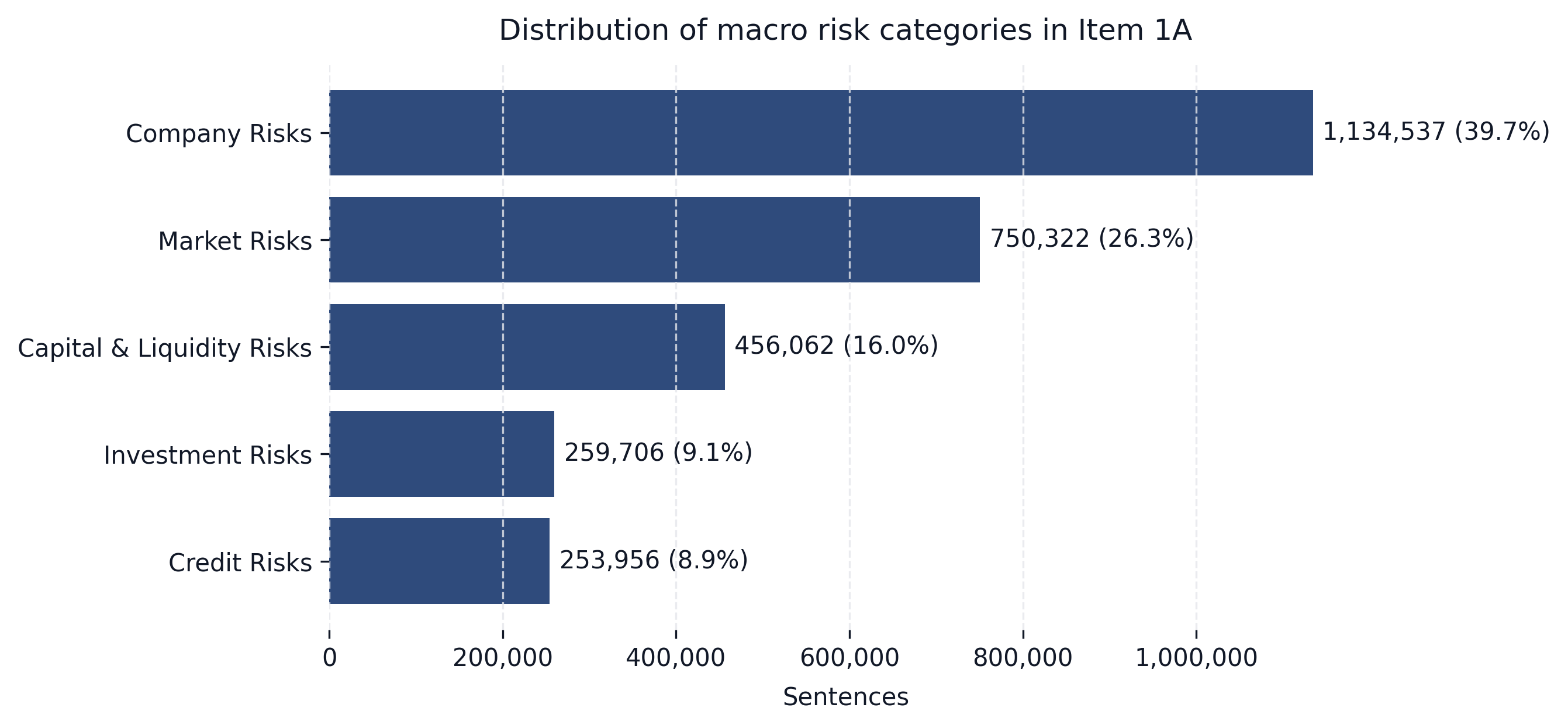}
\caption{Distribution of weakly-labeled risk sentences by \emph{macro} category (S\&P\,500, 2001--2025). Bars show counts; labels indicate count and share of all labels.}
\label{fig:macro-cat-prevalence}
\end{figure}

\begin{figure}[t]
\centering
\includegraphics[width=\linewidth]{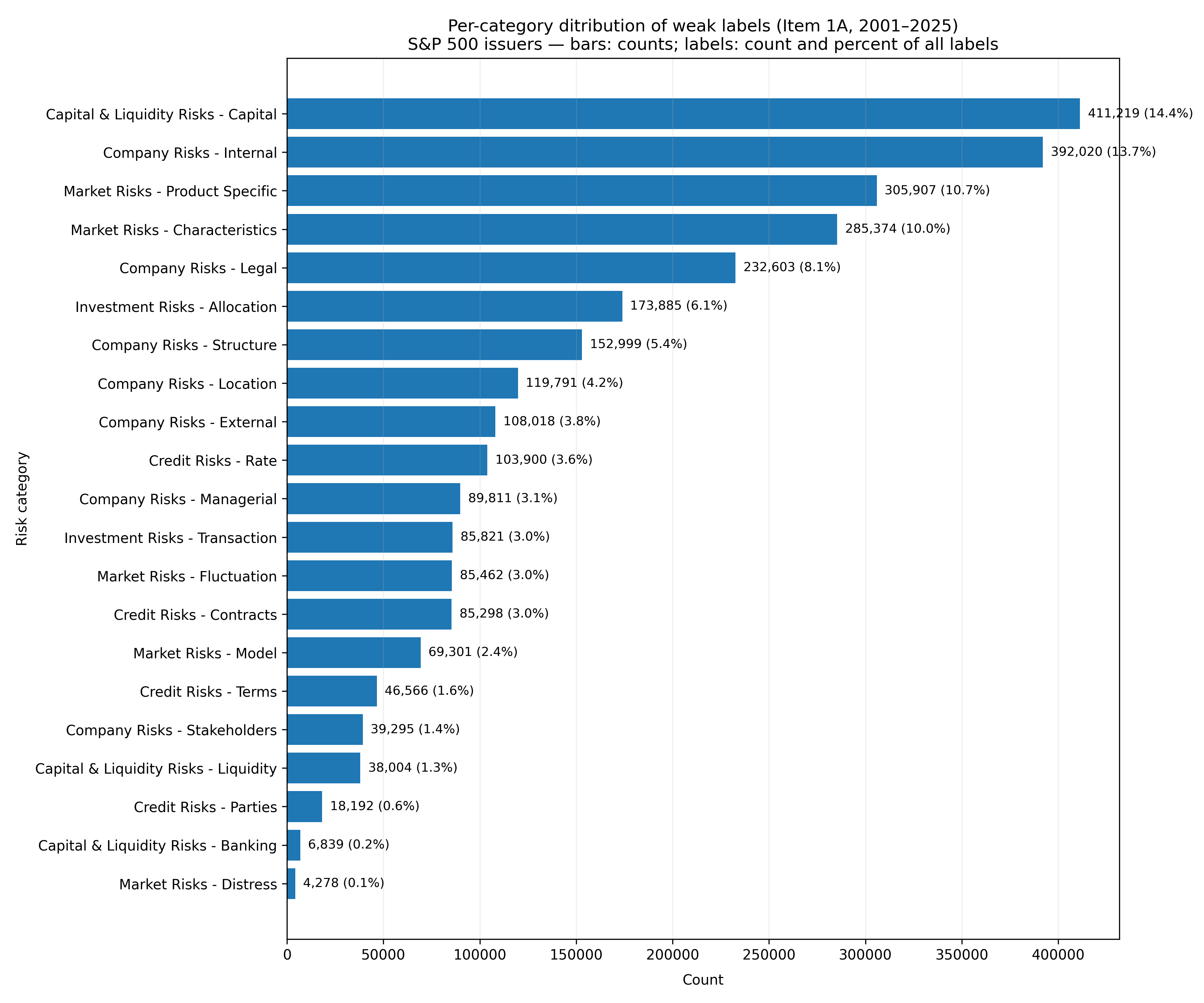}
\caption{Per-category distribution of weakly-labeled risk sentences (S\&P\,500, 2001--2025). Bars show counts; labels indicate count and share of all labels.}
\label{fig:cat-prevalence}
\end{figure}

\begin{figure}[t]
\centering
\includegraphics[width=\linewidth]{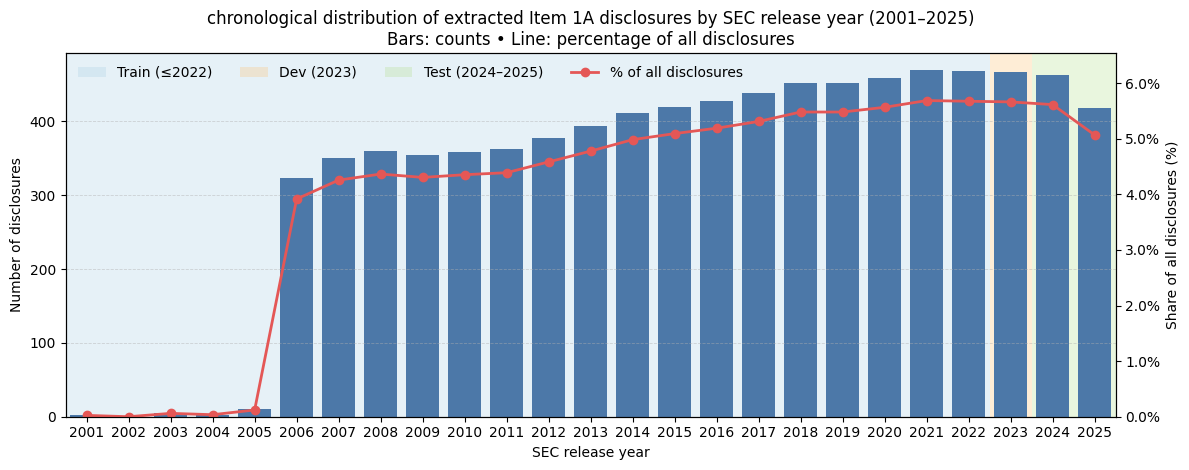}
\caption{Yearly distribution of extracted Item~1A \emph{disclosures} by SEC release year (2001--2025). Bars show counts; the line shows the percentage of all disclosures. Shaded bands indicate Train (\(\leq\)2022), Dev (2023), and Test (2024--2025).}
\label{fig:yearly-disclosures}
\end{figure}

\begin{figure}[t]
\centering
\includegraphics[width=\linewidth]{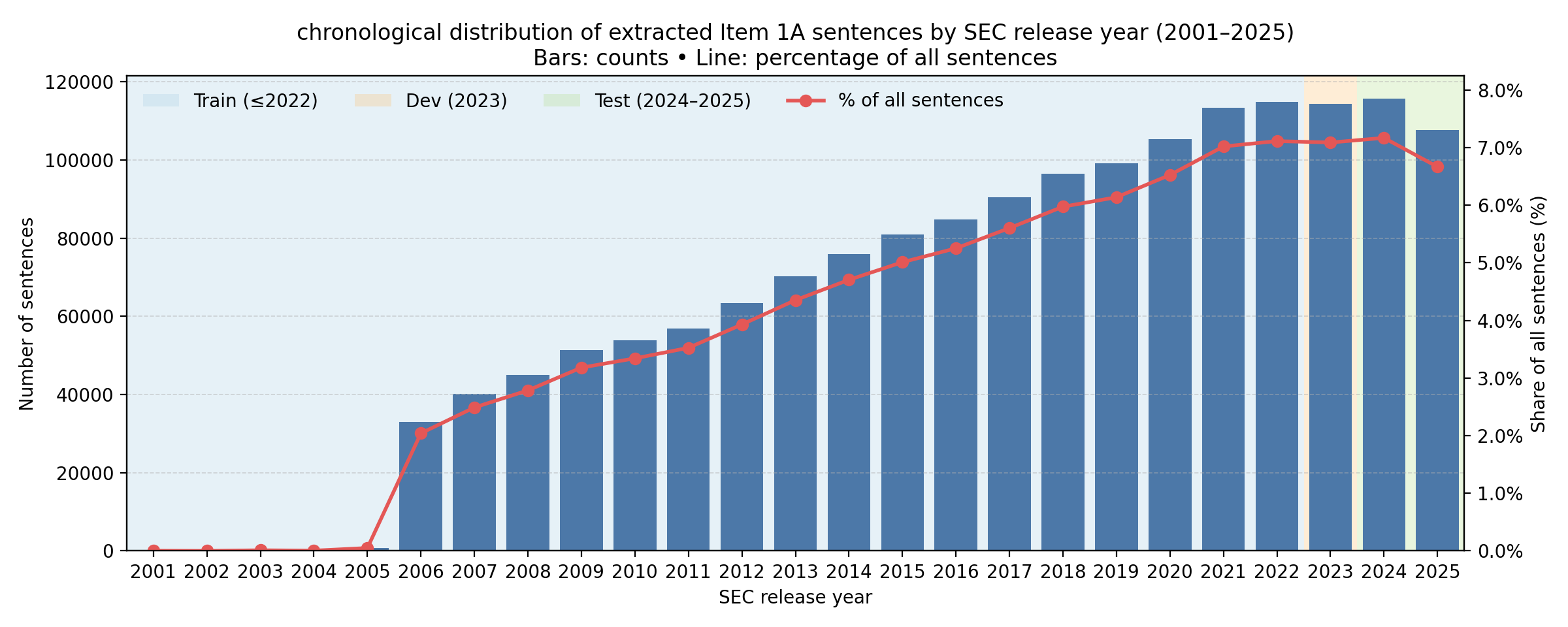}
\caption{Yearly distribution of extracted Item~1A \emph{sentences} by SEC release year (2001--2025). Bars show sentence counts; the line shows the percentage of all sentences. Shaded bands indicate Train (\(\leq\)2022), Dev (2023), and Test (2024--2025).}
\label{fig:yearly-sentences}
\end{figure}

To contextualize label balance and temporal coverage, we provide four summary figures: (i) a bar chart of \emph{macro} category prevalence aggregated across issuers (Fig.~\ref{fig:macro-cat-prevalence}); (ii) a bar chart of the 21 \emph{subcategory} labels (Fig.~\ref{fig:cat-prevalence}); (iii) yearly counts of Item~1A \emph{disclosures} by SEC release year (Fig.~\ref{fig:yearly-disclosures}); and (iv) yearly counts of \emph{sentences} by SEC release year (Fig.~\ref{fig:yearly-sentences}). The two timeline plots are shaded with the fixed split bands (Train \(\leq\)2022, Dev \(=\) 2023, Test \(=\) 2024--2025).

\end{document}